\documentclass[12pt]{article}
\usepackage[utf8]{inputenc}
\usepackage[a4paper, margin=2.5cm]{geometry}
\usepackage{graphicx}
\usepackage{amsmath}
\usepackage{natbib}
\usepackage{url}
\usepackage{hyperref}
\usepackage{subcaption}
\usepackage[font=small]{caption} 
\usepackage{placeins} 
\usepackage{tipa}

\hypersetup{
    colorlinks=true,
    linkcolor=blue,
    citecolor=blue,
    urlcolor=blue
}

\title{The 2D+ Dynamic Articulatory Model DYNARTmo:\\ Tongue-Palate Contact Area Estimation}
\author{Bernd J. Kröger\\
\small Medical Faculty, RWTH Aachen University, Aachen, Germany\\
\small Kröger Lab, Belgium, \url{www.speechtrainer.eu}}
\date{}

\begin{document}

\maketitle

\begin{abstract}
This paper describes an extension of the two-dimensional dynamic articulatory model DYNARTmo by integrating an 
internal three-dimensional representation of the palatal dome to estimate tongue–palate contact 
areas from midsagittal tongue contours. Two alternative dome geometries—a half-ellipse and 
a cosine-based profile—are implemented to model lateral curvature in the coronal plane. 
Using these geometries, lateral contact points are analytically computed for each anterior -posterior 
position, enabling the generation of electropalatography-like visualizations within the 2D+ framework. 
The enhanced model supports three synchronized views (sagittal, glottal, and palatal) for 
static and dynamic (animated) articulation displays, suitable for speech science education and 
speech therapy. Future work includes adding a facial (lip) view and implementing 
articulatory-to-acoustic synthesis to quantitatively evaluate model realism.
\end{abstract}

\section{Introduction}

In addition to midsagittal articulatory information, the estimation of the tongue-palate contact area during the 
production of consonants and non-low vowels provides essential insights into the geometry of articulation. This 
information is crucial not only for understanding the spatial configuration of the tongue but also for representing 
the proprioceptive and tactile feedback experienced by speakers during articulation.

Visualizations based on midsagittal views, such as x-ray or MRI data \citep{bressmann2005,narayanan2004}, and 
electropalatographic (EPG) recordings \citep{gibbon1999,hardcastle1991}, have long served as complementary methods 
for capturing both the internal structure of the vocal tract and surface-level contact patterns. While tube models 
based on 2D sagittal contours can approximate vocal tract area functions \citep{perrier1992}, the estimation remains 
imprecise without incorporating lateral or 3D structural information. Only full 3D reconstructions enable accurate 
computation of the vocal tract cross-sectional area, which is essential for realistic acoustic simulation 
\citep{takemoto2004, story1996}.

First evaluation studies in the context of speech therapy have shown that full 3D visualizations of articulatory 
movements can be too complex or cognitively demanding for patients with speech impairments. Instead, many patients 
and learners prefer simplified 2D sagittal animations, which offer a clearer representation of the main articulatory 
movements \citep{kroeger2013, hoole2012}. To strike a balance between simplicity and richness of information, 
we propose an enhanced 2D+ model that integrates midsagittal contours with calculated tongue-palate contact area 
patterns.

This paper builds upon the DYNARTmo model as described in \citet{kroeger2025arxiv}, with a specific focus on how 
to estimate tongue-palate contact using an internal three-dimensional model of the palatal dome. By integrating 
this internal 3D geometry, we enable the estimation of surface contact from the midsagittal contour of the tongue, 
resulting in EPG-like contact visualizations within a 2D framework.

\section{The 3D Model for Tongue-Palate Contact Estimation}
\subsection{Mathematical models for the palatal dome}

\begin{figure}[h]
  \centering
  \includegraphics[width=0.8\textwidth]{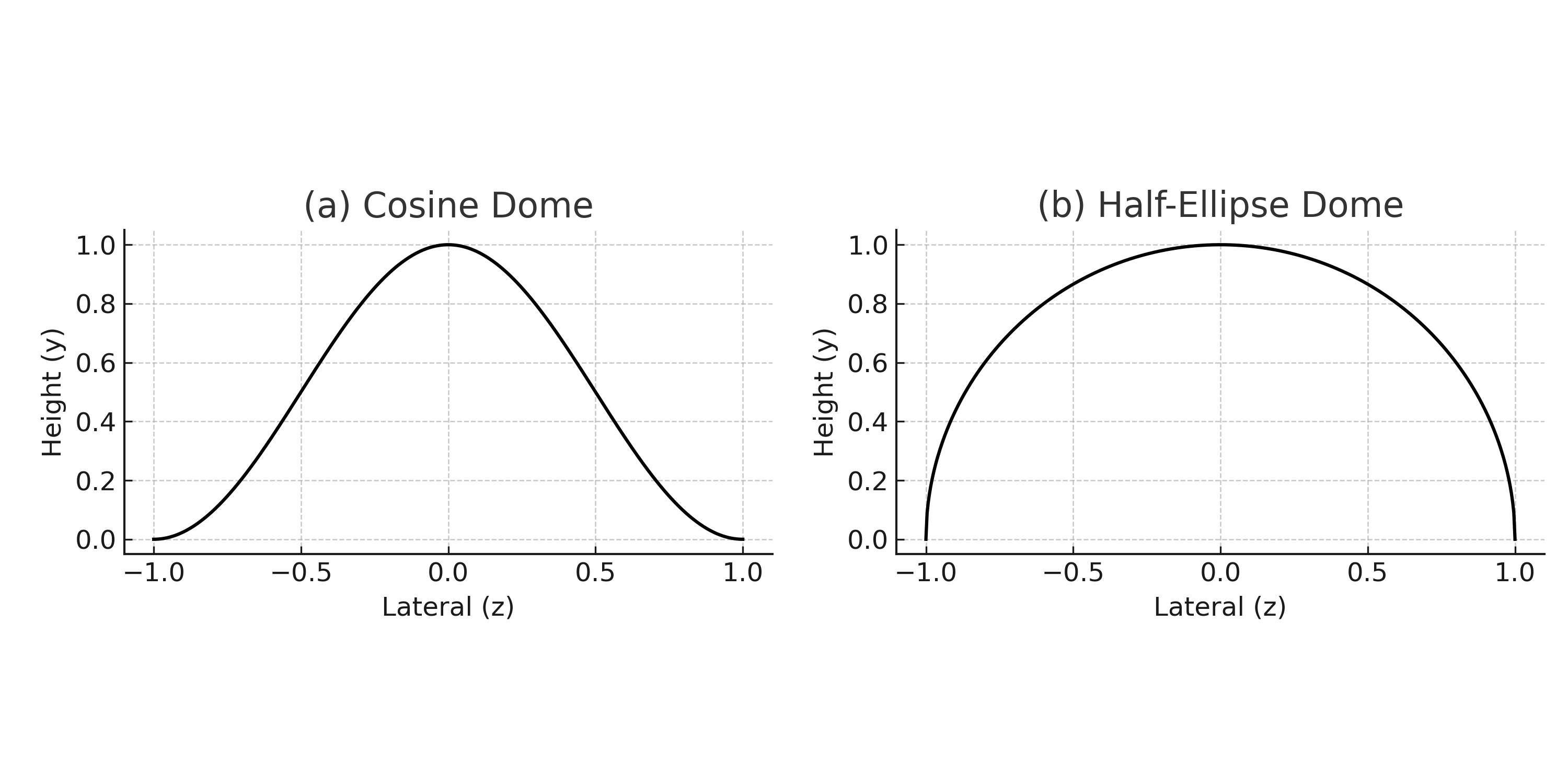}
  \caption{Geometrical representation of the lateral curvature of the palatal dome: (a) half-ellipse model and 
  (b) cosine dome model, both shown in the coronal (frontal) plane.}
  \label{fig:palatal_dome_shapes}
\end{figure}

Two mathematically distinct parametric models are proposed to approximate the lateral curvature of the palatal dome: 
a half-elliptical contour and a cosine-based dome shape (see Fig.~\ref{fig:palatal_dome_shapes}). 
These models describe the cross-sectional geometry of the hard palate in the coronal (frontal) plane at any 
given anterior-posterior (x) position.

The vertical dome height $h$ is measured from the baseline defined by the upper teeth row (i.e., from the alveolar 
ridge level). The lateral span at each anterior-posterior position is given by the distance between the left and 
right upper molars (denoted $z_l$ = -1 and $z_r$ = 1, respectively). Hence, each coronal slice of the dome model 
lies within the $y$–$z$ plane.

In the half-elliptical model, the dome rises steeply and symmetrically from the alveolar ridge, resulting in a narrow, 
high-arched shape (Fig.~\ref{fig:palatal_dome_shapes}a). In contrast, the cosine dome incorporates the gingival ridge 
behind the teeth, producing a smoother and anatomically more realistic curvature with a flatter onset 
(Fig.~\ref{fig:palatal_dome_shapes}b). This gradual curvature is especially important for modeling lateral 
tongue-palate contact during apical stops, nasals, and fricatives, where the tongue contacts the lateral sides 
of the hard palate just behind the alveolar ridge.

The mathematical formulation of the two models is given as follows:

\paragraph{Cosine Dome:}
\[
y(z) = y_\text{max} - \frac{h}{2} \left(1 - \cos\left(2\pi \cdot \frac{z - z_\text{min}}{z_\text{max} - z_\text{min}}\right)\right)
\]

\paragraph{Half-Elliptical Dome:}
\[
\left(\frac{z - z_\text{center}}{a}\right)^2 + \left(\frac{y - y_\text{max}}{b}\right)^2 = 1
\quad \text{for } y \leq y_\text{max}
\]

\begin{figure}[ht]
  \centering
  \begin{subfigure}[b]{0.48\textwidth}
    \includegraphics[width=\textwidth]{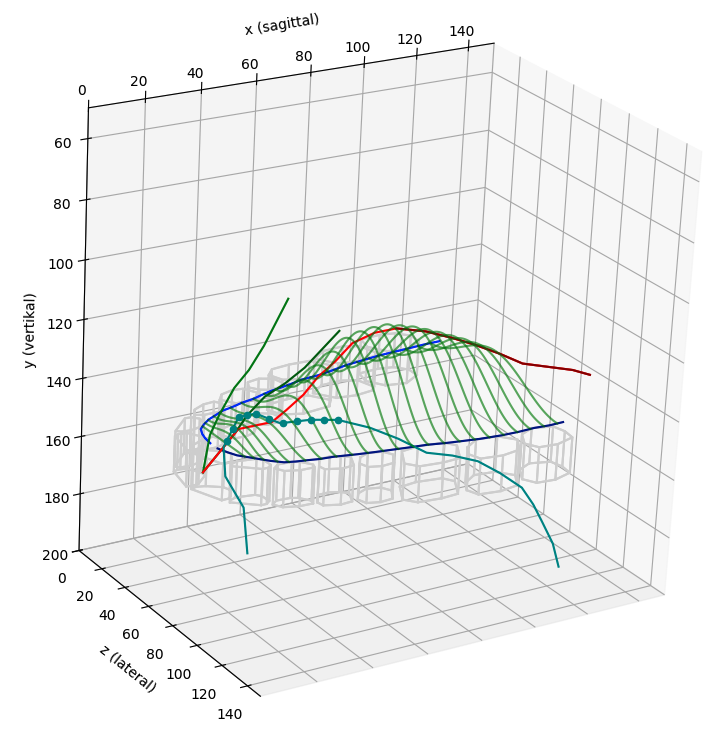}
    \caption{Cosine dome model (tongue forms a /t/)}
    \label{fig:fig2a}
  \end{subfigure}
  \hfill
  \begin{subfigure}[b]{0.48\textwidth}
    \includegraphics[width=\textwidth]{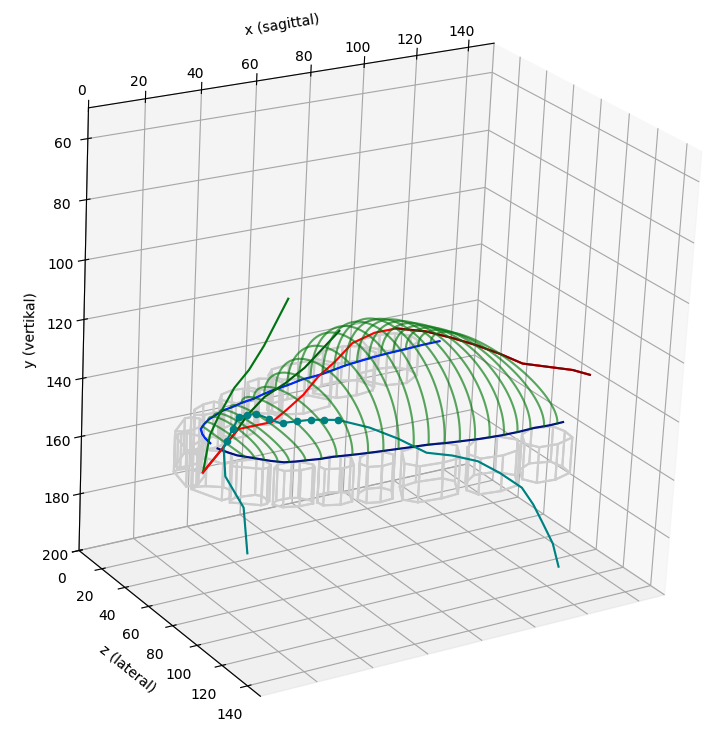}
    \caption{Half-ellipse model (tongue forms a /t/)}
    \label{fig:fig2b}
  \end{subfigure}
  \caption{3D reconstruction of the hard palate surface, including velar transition, using 
  (a) the cosine dome model and (b) the half-ellipse model. Coordinate system: $x$ denotes 
  anterior-to-posterior, $y$ inferior-to-superior, and $z$ medial-to-lateral (left and right).
  The tongue forms a /t/. Curves, lines and points: 
  the curves for palate dome in lateral plane (green); 
  the line of tongue (midsagittal) 
  including points for potential lateral tongue-palate contact (green); 
  midsagittal curve of palate (red); 
  lines of the gingival ridge (alveolar ridge) behind upper teeth row (blue).}
  \label{fig:palatal_dome_tt}
\end{figure}

\begin{figure}[ht]
  \centering
  \begin{subfigure}[b]{0.48\textwidth}
    \includegraphics[width=\textwidth]{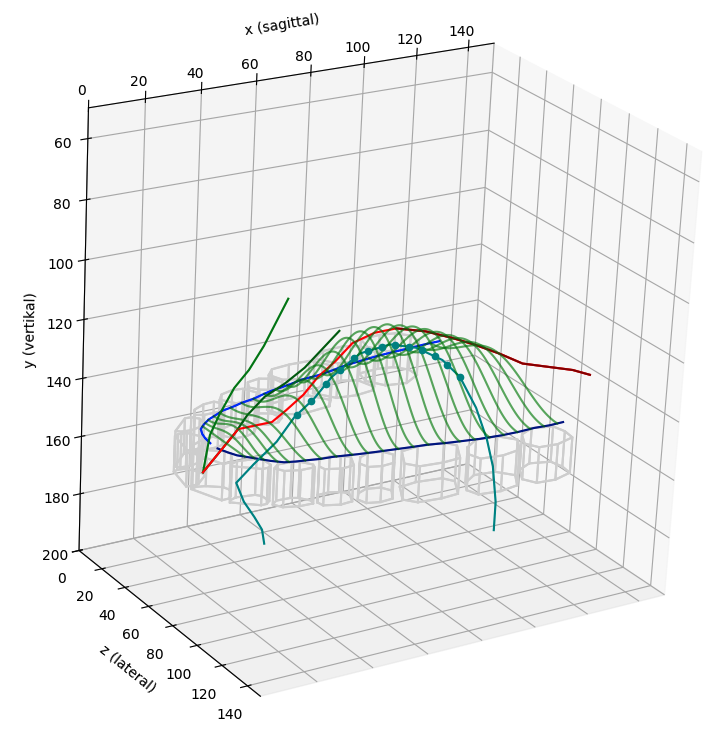}
    \caption{Cosine dome model (tongue forms an /i:/)}
    \label{fig:fig3a}
  \end{subfigure}
  \hfill
  \begin{subfigure}[b]{0.48\textwidth}
    \includegraphics[width=\textwidth]{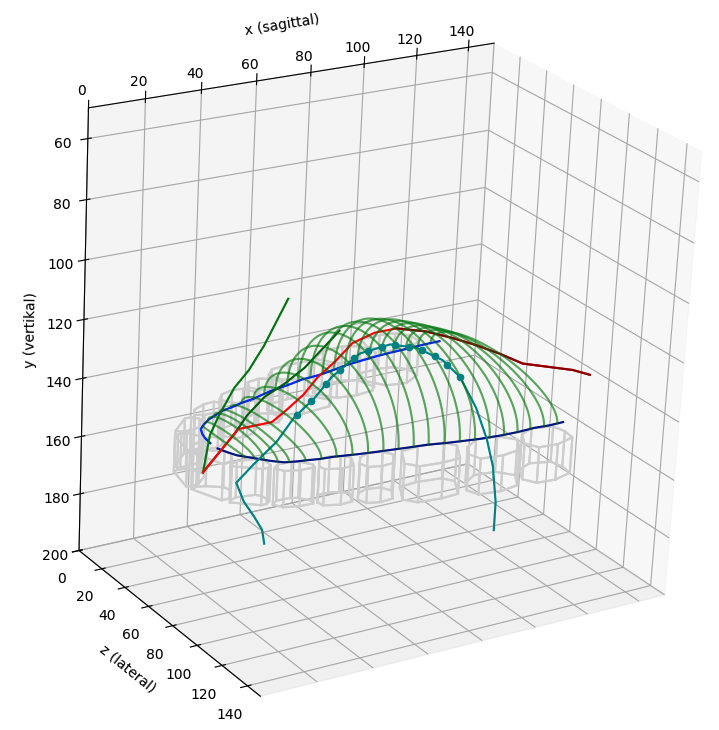}
    \caption{Half-ellipse model (tongue forms an /i:/)}
    \label{fig:fig3b}
  \end{subfigure}
  \caption{3D reconstruction of the hard palate surface, including velar transition, using 
  (a) the cosine dome model and (b) the half-ellipse model (cf. Figure 2).
  The tongue forms an /i:/.}
  \label{fig:palatal_dome_ii}
\end{figure}

To generate the full three-dimensional palatal surface, these lateral cross-sectional profiles are extended along 
the entire anterior-posterior axis (x-direction), covering the region from the incisors to the velar transition 
zone (see Fig.~\ref{fig:palatal_dome_tt} and see Fig.~\ref{fig:palatal_dome_ii}).

\subsection{A preliminary 3D tongue model and calculation of tongue-palate contact area}

Our 3D tongue model is simple but effective. For each anterior–posterior position $x$ along the 
midsagittal tongue contour, we assume that the lateral shape of the tongue in the coronal ($y$–$z$) 
plane can be approximated by a horizontal line at the corresponding tongue height $y$. 
This assumption holds for tongue regions located between the alveolar ridge (bottom of the dome) 
and the highest point of the palatal dome. Within this domain, the lateral intersection between 
the tongue and the palatal surface can be calculated analytically.

\paragraph{Cosine Dome – calculating $z$ for a given $y$:}
\[
\cos(2\pi t) = 1 - 2 \cdot \frac{y - y_\text{max}}{h}
\quad \Rightarrow \quad
t = \frac{1}{2\pi} \cdot \arccos\left(1 - 2 \cdot \frac{y - y_\text{max}}{h}\right)
\]
\[
z = z_\text{min} + t \cdot (z_\text{max} - z_\text{min})
\quad \text{and} \quad
z' = z_\text{max} - t \cdot (z_\text{max} - z_\text{min})
\]

\paragraph{Half-Elliptical Dome – calculating $z$ for a given $y$:}
\[
z = z_\text{center} \pm a \cdot \sqrt{1 - \left(\frac{y - y_\text{max}}{b}\right)^2}
\]

with: $a = w$ (half of the lateral dome width) and  $b = h$ (dome height).

\begin{figure}[ht]
  \centering
  \begin{subfigure}[b]{0.48\textwidth}
    \includegraphics[width=\textwidth]{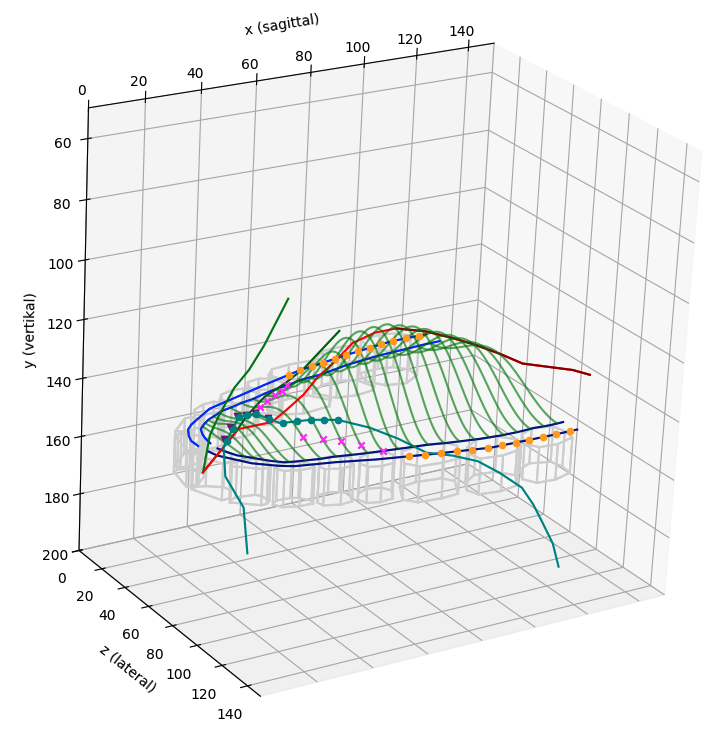}
    \caption{Cosine dome model}
  \end{subfigure}
  \hfill
  \begin{subfigure}[b]{0.48\textwidth}
    \includegraphics[width=\textwidth]{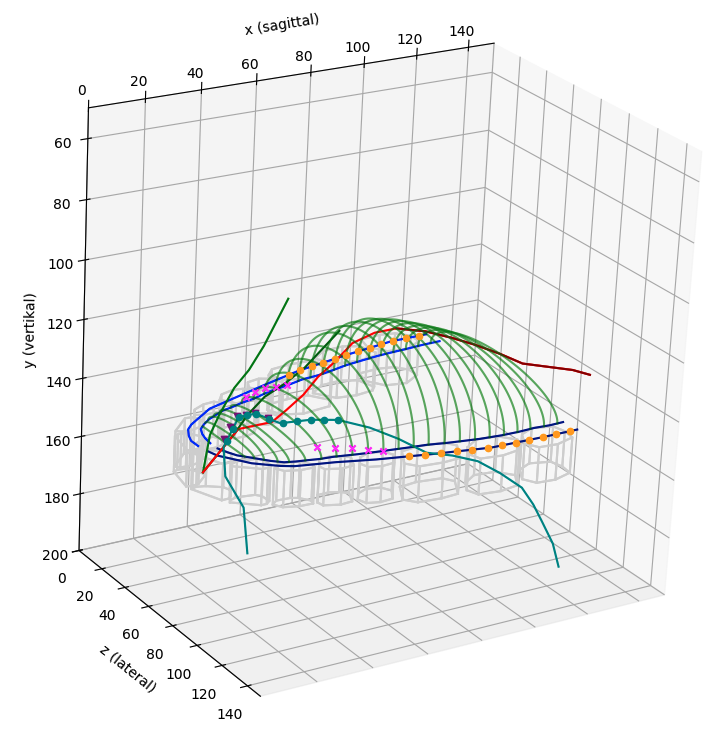}
    \caption{Half-ellipse model}
  \end{subfigure}
  \caption{3D reconstruction of the hard palate with velar transition during /t/-production. 
  Tongue-palate contact points are color-coded: yellow = no contact, red crosses = intersection, 
  full contact = central touchpoint. 
  Tongue-palate contact points are color-coded: yellow = no contact, red crosses = inter-section, 
  full contact = central touchpoint.}
  \label{fig:fig4}
\end{figure}

\begin{figure}[ht]
  \centering
  \begin{subfigure}[b]{0.48\textwidth}
    \includegraphics[width=\textwidth]{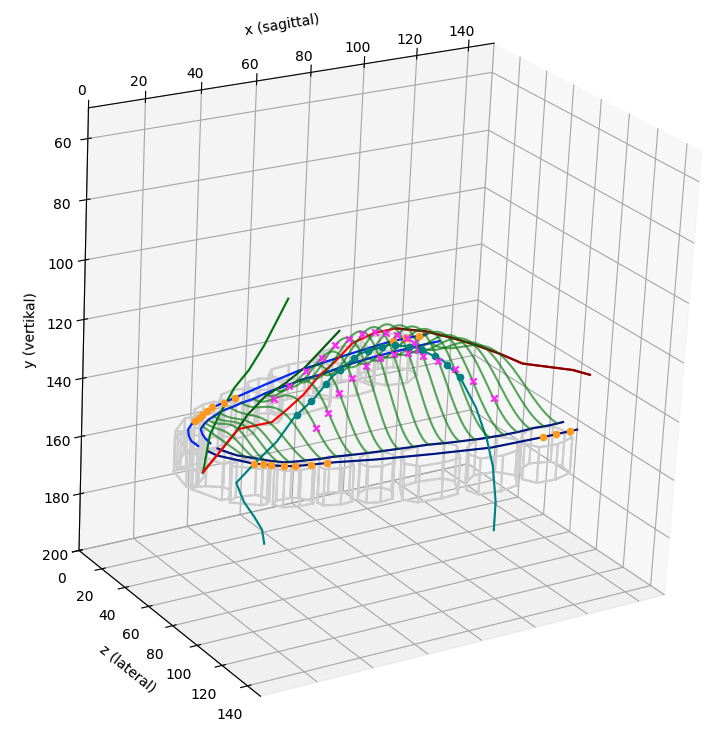}
    \caption{Cosine dome model}
  \end{subfigure}
  \hfill
  \begin{subfigure}[b]{0.48\textwidth}
    \includegraphics[width=\textwidth]{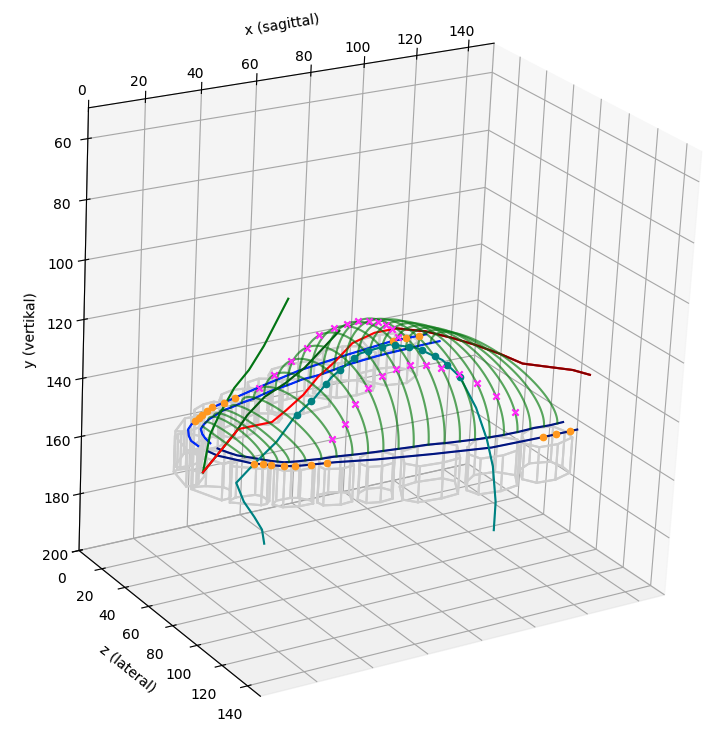}
    \caption{Half-ellipse model}
  \end{subfigure}
  \caption{Same as Figure~\ref{fig:fig4}, but for the vowel /i:/.}
  \label{fig:fig5}
\end{figure}

But even other cases may occur: In total, the following three geometric cases must be distinguished at 
the arch of dome $x$-positions:

\begin{enumerate}
  \item \textbf{No contact:} If the tongue is located below the alveolar ridge (i.e., $y$ is smaller than 
  the lower boundary of the dome), no intersection can be computed. These regions are indicated by yellow 
  points on the teeth row in Figures~\ref{fig:fig4} and~\ref{fig:fig5} (see, e.g., /t/ and /i:/).
  
  \item \textbf{Full contact:} If the tongue lies entirely above the dome surface (e.g., anterior closure 
  in /t/), a full lateral tongue-palate contact is assumed across the entire dome span at that $x$-position. 
  he corresponding intersection is set to the peak of the dome 
  (i.e., $y = y_\text{max}$, $z = z_\text{center}$ or $z = (z_\text{min} + z_\text{max}) / 2$). 
  See the anterior region in Figure~\ref{fig:fig4}.

  \item \textbf{Intersection exists:} In all other cases, the horizontal tongue level intersects the dome 
  shape at two lateral points (calculated via the formulas above). These points correspond to the 
  tongue-palate contact locations in the left and right lateral regions. They are marked with red crosses 
  in Figures~\ref{fig:fig4} and~\ref{fig:fig5}.
\end{enumerate}

The resulting contact area patterns for the sounds /t/ and /i:/ as well as for other speech sounds
are displayed by DYNARTmo - which now includes the 3D model for tongue-palate contact area 
estimation - and are given here in the next section of this paper.

\FloatBarrier

\section{Integration of the 3D Model into DYNARTmo}
\subsection{Control parameters and lateral articulator shapes}

\begin{figure}[ht]
  \centering

  \begin{subfigure}[b]{0.22\textwidth}
    \includegraphics[width=\textwidth]{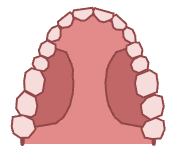}
    \caption{sound /i:/}
  \end{subfigure}
  \hfill
  \begin{subfigure}[b]{0.22\textwidth}
    \includegraphics[width=\textwidth]{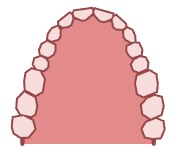}
    \caption{sound /a:/}
  \end{subfigure}
  \hfill
  \begin{subfigure}[b]{0.22\textwidth}
    \includegraphics[width=\textwidth]{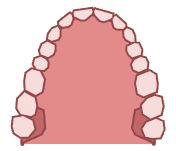}
    \caption{sound /u:/}
  \end{subfigure}
  \hfill
  \begin{subfigure}[b]{0.22\textwidth}
    \includegraphics[width=\textwidth]{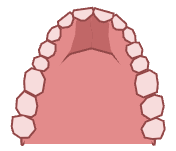}
    \caption{sound /l/}
  \end{subfigure}

  \vspace{0.3cm}

  \begin{subfigure}[b]{0.22\textwidth}
    \includegraphics[width=\textwidth]{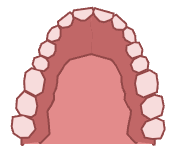}
    \caption{sound /t/}
  \end{subfigure}
  \hfill
  \begin{subfigure}[b]{0.22\textwidth}
    \includegraphics[width=\textwidth]{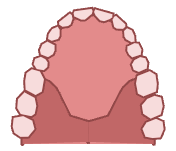}
    \caption{sound /k/}
  \end{subfigure}
  \hfill
  \begin{subfigure}[b]{0.22\textwidth}
    \includegraphics[width=\textwidth]{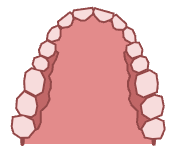}
    \caption{sound /\textipa{T}/}
  \end{subfigure}
  \hfill
  \begin{subfigure}[b]{0.22\textwidth}
    \includegraphics[width=\textwidth]{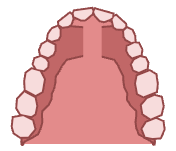}
    \caption{sound /s/}
  \end{subfigure}

  \vspace{0.3cm}

  \begin{subfigure}[b]{0.22\textwidth}
    \includegraphics[width=\textwidth]{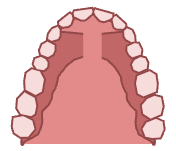}
    \caption{sound /\textipa{S}/} 
  \end{subfigure}
  \hfill
  \begin{subfigure}[b]{0.22\textwidth}
    \includegraphics[width=\textwidth]{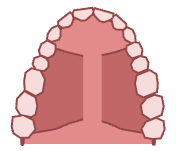}
    \caption{sound /\textipa{\textctc}/}
  \end{subfigure}
  \hfill
  \begin{subfigure}[b]{0.22\textwidth}
    \includegraphics[width=\textwidth]{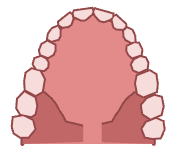}
    \caption{sound /x/}
  \end{subfigure}
  \hfill
  \begin{subfigure}[b]{0.22\textwidth}
    \includegraphics[width=\textwidth]{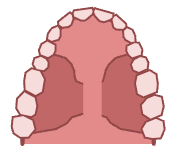}
    \caption{sound /j/}
  \end{subfigure}

  \caption{Tongue-palate contact patterns for vowels and consonants (cosine model).}
  \label{fig:fig6_tp_cosine}
\end{figure}

\begin{figure}[ht]
  \centering

  \begin{subfigure}[b]{0.22\textwidth}
    \includegraphics[width=\textwidth]{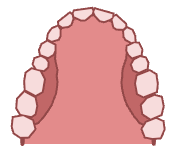}
    \caption{sound /i:/}
  \end{subfigure}
  \hfill
  \begin{subfigure}[b]{0.22\textwidth}
    \includegraphics[width=\textwidth]{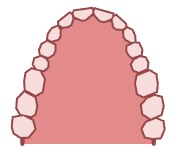}
    \caption{sound /a:/}
  \end{subfigure}
  \hfill
  \begin{subfigure}[b]{0.22\textwidth}
    \includegraphics[width=\textwidth]{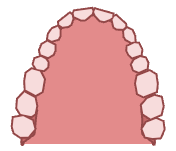}
    \caption{sound /u:/}
  \end{subfigure}
  \hfill
  \begin{subfigure}[b]{0.22\textwidth}
    \includegraphics[width=\textwidth]{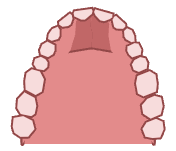}
    \caption{sound /l/}
  \end{subfigure}

  \vspace{0.3cm}

  \begin{subfigure}[b]{0.22\textwidth}
    \includegraphics[width=\textwidth]{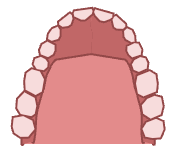}
    \caption{sound /t/}
  \end{subfigure}
  \hfill
  \begin{subfigure}[b]{0.22\textwidth}
    \includegraphics[width=\textwidth]{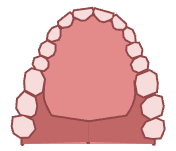}
    \caption{sound /k/}
  \end{subfigure}
  \hfill
  \begin{subfigure}[b]{0.22\textwidth}
    \includegraphics[width=\textwidth]{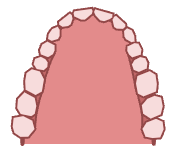}
    \caption{sound /\textipa{T}/}
  \end{subfigure}
  \hfill
  \begin{subfigure}[b]{0.22\textwidth}
    \includegraphics[width=\textwidth]{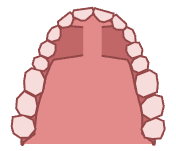}
    \caption{sound /s/}
  \end{subfigure}

  \vspace{0.3cm}

  \begin{subfigure}[b]{0.22\textwidth}
    \includegraphics[width=\textwidth]{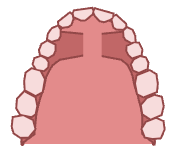}
    \caption{sound /\textipa{S}/}
  \end{subfigure}
  \hfill
  \begin{subfigure}[b]{0.22\textwidth}
    \includegraphics[width=\textwidth]{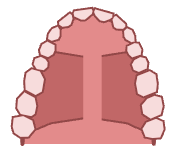}
    \caption{sound /\textipa{\textctc}/}
  \end{subfigure}
  \hfill
  \begin{subfigure}[b]{0.22\textwidth}
    \includegraphics[width=\textwidth]{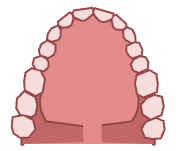}
    \caption{sound /x/}
  \end{subfigure}
  \hfill
  \begin{subfigure}[b]{0.22\textwidth}
    \includegraphics[width=\textwidth]{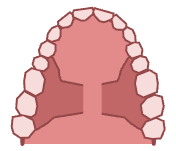}
    \caption{sound /j/}
  \end{subfigure}

  \caption{Tongue-palate contact patterns for vowels and consonants (half-ellipse model).}
  \label{fig:fig7_tp_ellipse}
\end{figure}

The integration of the “internal” 3D model for tongue–palate contact pattern generation, as described 
in Section~2, into DYNARTmo \citep{kroeger2025arxiv} enables the generation of tongue–palate contact 
patterns for vowels. However, in order to reproduce the full range of contact patterns for 
consonants (see Figures~\ref{fig:fig6_tp_cosine} and~\ref{fig:fig7_tp_ellipse}), 
it is necessary to model different shapes of the constriction-forming articulator—typically the 
tongue tip or the tongue dorsum in the case of most non-labial consonants—depending on 
the \textit{manner of articulation}. The parameter \textit{manner of articulation} is defined in the 
model for different articulators (see Table~2 in \citealp{kroeger2025arxiv}).

In DYNARTmo, changes in the shape of the tongue tip or tongue dorsum are controlled by the 
articulatory parameters \textit{tongue tip manner} and \textit{tongue dorsum manner}. 
For the \textit{tongue tip} parameter, possible discrete settings are \textit{full}, \textit{near}, 
and \textit{lateral}. For the \textit{tongue dorsum} parameter, only the states \textit{full} 
and \textit{near} are possible.

In the \textit{full} mode, consonantal articulation—unlike vowel articulation—requires an 
additional elevation of the lateral tongue edges in the posterior region for apical plosives, 
nasals, and fricatives. This prevents lateral leakage of the airstream and thus directs the 
airflow towards the apical constriction (in the case of fricatives) or stops the airflow within 
the oral cavity during the apical closure (in the case of plosives and nasals). In other words, 
in addition to the primary apical constriction, \textit{full} mode for these sounds includes a 
posterior dorsal elevation of the tongue edges.

\begin{figure}[ht]
  \centering
  \includegraphics[width=0.85\textwidth]{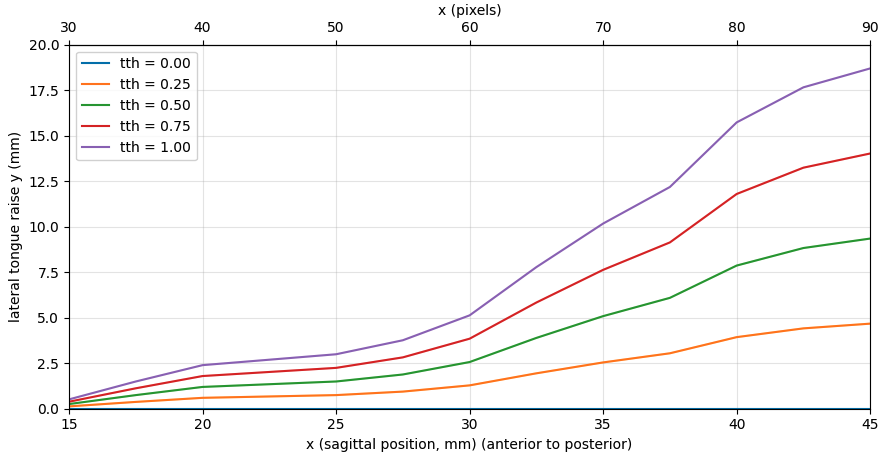}
  \caption{Degree of lateral tongue-edge elevation ($y$ in mm) as a function 
  of $x$ (anterior–posterior direction) and of the control parameter 
  \textit{tongue tip height} (tth) for apical plosives and fricatives.}
  \label{fig:fig8_lat_tongue_elevation}
\end{figure}

This lateral tongue-edge elevation is determined heuristically for each anterior–posterior ($x$) 
location along the tongue and is applied gradually as a function of the control parameter 
\textit{tongue tip height (tth)} (for definition of tth see Table~1 
in \citealp{kroeger2025arxiv}). The heuristic criterion 
is that the elevated lateral tongue edges make contact with the alveolar ridge (gingival ridge) 
even in the region of the molars. The resulting relationship is shown in 
Figure~\ref{fig:fig8_lat_tongue_elevation}.

Other tongue shapes that need to be implemented, such as \textit{lateral lowering} of the 
tongue for the lateral /l/ or \textit{central groove forming} of the tongue for all apical 
and dorsal fricatives, are currently realized in the model in a simplified way. 
Specifically, \textit{central groove forming} is modeled as a complete lowering of the 
central tongue region across a constant width over the entire tongue length (anterior to posterior), 
while \textit{lateral lowering} for the lateral /l/ is modeled as a complete lowering of the 
lateral tongue edges across a constant lateral width along the entire tongue length 
(see Figures~\ref{fig:fig6_tp_cosine} and~\ref{fig:fig7_tp_ellipse}).

\FloatBarrier

\subsection{Views: sagittal, glottal, palatal}

The primary application of DYNARTmo is the generation of visual information to facilitate the 
understanding, learning, or re-learning of articulation. This is aimed at students of 
linguistics, phonetics, and speech–language therapy, as well as at individuals suffering 
from different types of speech disorders.

\begin{figure}[ht]
  \centering
  \begin{subfigure}[b]{0.48\textwidth}
    \includegraphics[width=\textwidth]{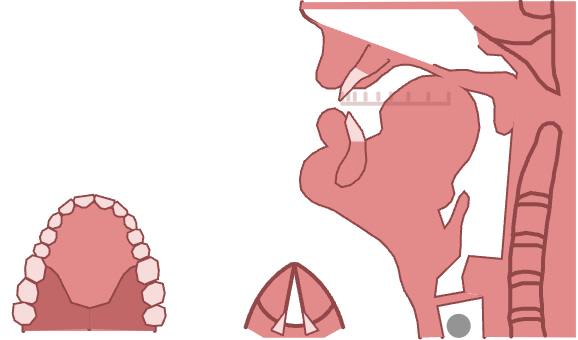}
    \caption{/k/ during closure phase}
  \end{subfigure}
  \hfill
  \begin{subfigure}[b]{0.48\textwidth}
    \includegraphics[width=\textwidth]{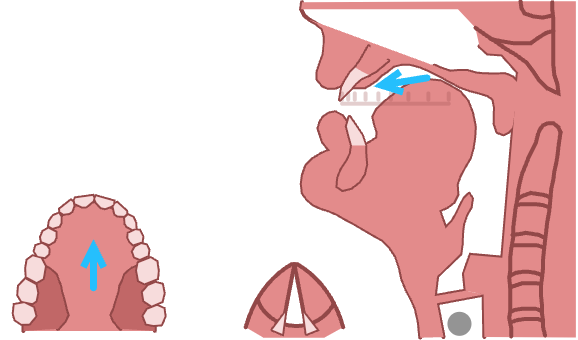}
    \caption{/k/ shortly after release}
  \end{subfigure}

  \vspace{0.3cm}

  \begin{subfigure}[b]{0.48\textwidth}
    \includegraphics[width=\textwidth]{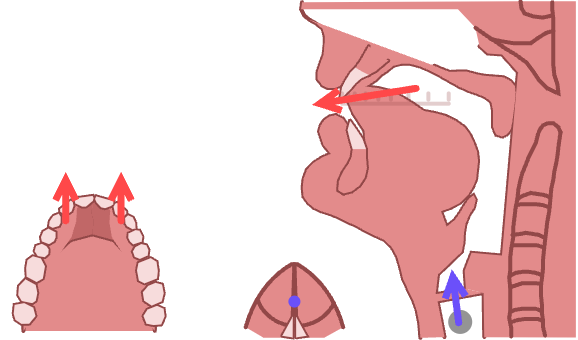}
    \caption{/l/ (apical lateral)}
  \end{subfigure}
  \hfill
  \begin{subfigure}[b]{0.48\textwidth}
    \includegraphics[width=\textwidth]{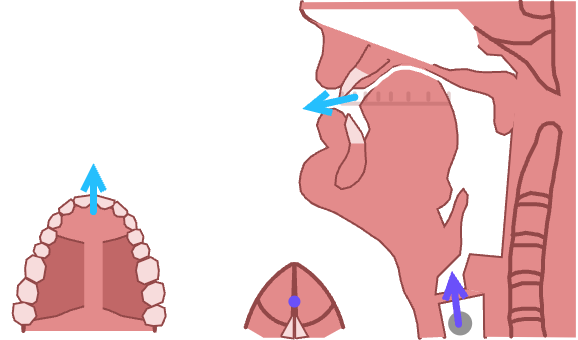}
    \caption{/\textipa{J}/ (voiced palatal fricative)}
  \end{subfigure}

  \caption{Sagittal, glottal, and palatal views generated by DYNARTmo. Top row: /k/ during 
  closure and shortly after release; bottom row: /l/ (apical lateral) and /\textipa{J}/ 
  (voiced palatal fricative).}
  \label{fig:fig9_dynartmo_views}
\end{figure}

Since the acoustically and phonologically crucial distinction between voiced and voiceless 
sounds cannot be determined from the midsagittal view alone, we have added a \textit{glottal view} 
in addition to the \textit{sagittal view}—even though this may initially seem confusing for 
some patients. This addition prominently visualizes the difference between voiced and voiceless 
sounds: an open glottis (breathing position) for voiceless sounds versus a closed or loosely 
closed glottis for phonation in voiced sounds, accompanied by the corresponding movement of 
the arytenoid cartilages (see Figure~\ref{fig:fig9_dynartmo_views}).

In addition to the sagittal and glottal views, DYNARTmo now also includes a \textit{palatal view} 
for visualizing the tongue–palate contact area and contact pattern whenever the tongue touches 
the palate. This view enables learners or patients to link visual information to their tactile 
perception of the tongue-palate contact. While the midsagittal (sagittal) view primarily supports 
awareness of proprioceptive feedback related to the positioning of articulators via tongue 
musculature, jaw musculature, and temporomandibular joint movement, the palatal view emphasizes 
tactile sensation of tongue–palate contact.

The visualizations also indicate the location of the primary and secondary acoustic sources of 
speech sounds. The primary source—phonation at the glottis for voiced sounds—is marked by a 
violet arrow in the sagittal view and a violet dot in the glottal view
(e.g., Figure~\ref{fig:fig9_dynartmo_views}c or d). The secondary source—turbulent 
airflow generated downstream from an oral constriction—is marked by a light blue arrow. 
A turbulence arrow is displayed not only for fricatives (e.g., Figure~\ref{fig:fig9_dynartmo_views}d) 
but also for plosives at the moment of release, or for a short time thereafter, to visualize the 
noise burst (e.g., Figure~\ref{fig:fig9_dynartmo_views}b). An orange arrow is used to mark 
non-turbulent airflow, such as the lateral airstream in laterals (e.g., 
Figure~\ref{fig:fig9_dynartmo_views}c). Furthermore, the required subglottal pressure is 
indicated by a grey dot of varying size in the trachea, below the glottis.

These aerodynamic indicators (arrows and dots) are implemented not only in the static DYNARTmo 
images but also in the video animations of articulatory movements for syllables and words, 
displayed during the relevant time intervals.

\section{Discussion and Further Work}

DYNARTmo generates static articulatory snapshots and additionally allows the creation of animations 
(videos) of articulatory movements for syllables and words. The three visual representations—the 
sagittal, glottal, and palatal views—qualitatively depict the facts of speech articulation in a 
satisfactory and accurate way. 

A quantitative evaluation of the model is planned for the future by extending DYNARTmo with an 
articulatory–acoustic module. This will enable the generation of an acoustic speech signal 
directly from the geometric data underlying the model and from the modules responsible for 
generating speech sound targets and articulatory movements. Since the human ear is considerably 
more sensitive than the eye in assessing the quality of the articulatory geometries (speech sound 
targets) and the underlying articulatory movements (see, e.g., the motor theory of speech 
perception \citep{liberman1967, liberman1996, liberman1985, galantucci2006}), producing an intelligible and 
possibly even natural-sounding acoustic signal 
would provide strong evidence that the model generates spatio–temporal patterns which are 
quantitatively realistic.

In the next two steps, we plan to (i) extend DYNARTmo with a \textit{facial view} that displays 
the mouth region and thus lip movements, and (ii) implement a first approach to articulatory 
resynthesis, enabling the timing of DYNARTmo-generated articulatory movements to be aligned 
with the timing of given natural speech signals.

\section{Supplementary Material}

The 3D module for calculating tongue–palate contact patterns is available for download. This 
module is written in Python and is located in the folder \texttt{/py\_code\_3D\_module/}. 
The program can be executed in any Python interpreter by running the file \texttt{main.py}.

The following instructions are also provided in a \texttt{README.txt} file:

\begin{itemize}
    \item In \texttt{lib/input\_data.py}, the user can choose between the tongue contour for /i:/ or /t/.
    \item In \texttt{config\_palate\_shape.py}, the user can select either the cosine model or the half-elliptical model.
    \item The program can directly generate the results shown in Figures~\ref{fig:fig4} and~\ref{fig:fig5}, i.e., exactly those figures can be reproduced.
\end{itemize}

\bibliographystyle{apalike}
\bibliography{dynartmo_palate}

\end{document}